Pawan Kumar Singh[1], Supratim Das[2], Ram Sarkar[3] and Mita Nasipuri[4]

# A New Approach for Texture based Script Identification At Block Level using Quad-Tree Decomposition

**Abstract:** A considerable amount of success has been achieved in developing monolingual OCR systems for *Indic* scripts. But in a country like India, where multi-script scenario is prevalent, identifying scripts beforehand becomes obligatory. In this paper, we present the significance of Gabor wavelets filters in extracting directional energy and entropy distributions for 11 official handwritten scripts *namely*, *Bangla*, *Devanagari*, *Gujarati*, *Gurumukhi*, *Kannada*, *Malayalam*, *Oriya*, *Tamil*, *Telugu*, *Urdu* and *Roman.* The experimentation is conducted at block level based on a quad-tree decomposition approach and evaluated using six different well-known classifiers. Finally, the best identification accuracy of 96.86% has been achieved by Multi Layer Perceptron (MLP) classifier for 3-fold cross validation at level-2 decomposition. The results serve to establish the efficacy of the present approach to the classification of handwritten *Indic* scripts.

## 1 Introduction

Script is defined as the graphic form of writing system which is used to express the written languages. Languages throughout the world are typeset in many different scripts. A script may be used by only one language or shared by many

**1** Department of Computer Science and Engineering, Jadavpur University, Kolkata, India
pawansingh.ju@gmail.com
**2** Department of Computer Science and Engineering, Jadavpur University, Kolkata, India
supratimdas21@gmail.com
**3** Department of Computer Science and Engineering, Jadavpur University, Kolkata, India
raamsarkar@gmail.com
**4** Department of Computer Science and Engineering, Jadavpur University, Kolkata, India
mitanasipuri@gmail.com

languages, with slight variations from one language to other. For example, *Devanagari* is used for writing a number of Indian languages like *Hindi*, *Konkani*, *Sanskrit*, *Nepali*, etc., whereas *Assamese* and *Bengali* languages use different variants of the *Bangla* script. India is a multilingual country with 23 constitutionally recognized languages written in 12 major scripts. Besides these, hundreds of other languages are used in India, each one with a number of dialects. The officially recognized languages are *Hindi*, *Bengali*, *Punjabi*, *Marathi*, *Gujarati*, *Oriya*, *Sindhi*, *Assamese*, *Nepali*, *Urdu*, *Sanskrit*, *Tamil*, *Telugu*, *Kannada*, *Malayalam*, *Kashmiri*, *Manipuri*, *Konkani*, *Maithali*, *Santhali*, *Bodo*, *Dogari* and *English*. The 12 major scripts used to write these languages are: *Devanagari*, *Bangla*, *Oriya*, *Gujarati*, *Gurumukhi*, *Tamil*, *Telugu*, *Kannada*, *Malayalam*, *Manipuri*, *Roman* and *Urdu*. Of these, *Urdu* is derived from the *Persian* script and is written from right to left. The first 10 scripts are originated from the early *Brahmi* script (300 BC) and are also referred to as *Indic* scripts [1-2]. *Indic* scripts are a logical composition of individual script symbols and follow a common logical structure. This can be referred to as the "script composition grammar" which has no counterpart in any other scripts in the world. *Indic* scripts are written syllabically and are visually composed in three tiers where constituent symbols in each tier play specific roles in the interpretation of that syllable [1].

Script identification aims to extract information presented in digital documents *viz.*, articles, newspapers, magazines and e-books. Automatic script identification facilitates sorting, searching, indexing, and retrieving of multilingual documents. Each script has its own set of characters which is very different from other scripts. In addition, a single OCR system can recognize only a script of particular type. On the contrary, it is perhaps impossible to design a single recognizer which can identify a variety of scripts/languages. Hence, in this multilingual and multi-script environment, OCR systems are supposed to be capable of recognizing characters irrespective of the script in which they are written. In general, recognition of different script characters in a single OCR module is difficult. This is because of features which are necessary for character recognition depends on the structural property, style and nature of writing which generally differ from one script to another. Alternative option for handling documents in a multi-script environment is to use a pool of OCRs (different OCR for different script) corresponding to different scripts. The characters in an input document can then be recognized reliably by selecting the appropriate OCR system from the said pool. However, it requires a priori knowledge of the script in which the document is written. Unfortunately, this information may not be readily available. At the same time, manual identification of the documents' scripts may be monotonous and time consuming. Therefore, it is necessary to identify the script of the document before feeding the document to the corresponding OCR system.

In general, script identification can be achieved at any of the three levels: (a) Page level, (b) Text-line level and (c) Word level. Identifying scripts at page-level can be sometimes too convoluted and protracted. Again, the identification of scripts at text-line or word level requires the exact segmentation of the document pages into their corresponding text lines and words. That is, the accuracy of the script identification in turns depends on the accuracies of its text-line and word segmentation methods respectively. In addition, identifying text words of different scripts with only a few numbers of characters may not always be feasible because at word-level, the number of characters present in a single word may not be always informative. So, in the present work, we have performed script identification at block level, where the entire script document pages are decomposed gradually into its corresponding smaller blocks using a quad-tree based segmentation approach.

The literature survey on the present topic reveals that very few researchers had performed script identification at block level other than three accustomed levels. P. K. Singh *et al.* [3] presented a detailed survey of the techniques applied on both printed and handwritten *Indic* script identification. M. Hangarge *et al.* [4] proposed a handwritten script identification scheme to identify 3 different scripts *namely*, *English*, *Devanagari*, and *Urdu*. A set of 13 spatial spread features were extracted using morphological filters. Further, $k$- Nearest Neighbor ($k$-NN) algorithm was used to classify 300 text blocks attaining an accuracy rate of 88.6% with five fold cross validation test. G.G. Rajput *et al.* [5] proposed a novel method towards multi-script identification at block level on 8 handwritten scripts including *Roman*. The recognition was based upon features extracted using Discrete Cosine Transform (DCT) and Wavelets of Daubechies family. Identification of the script was done using $k$-NN classifier on a dataset of 800 text block images and yielded an average recognition rate of 96.4%. G. D. Joshi *et al.* [6] presented a scheme to identify 10 different printed *Indic* scripts. The features were extracted globally from the responses of a multi-channel log-Gabor filter bank. This proposed system achieved an overall classification accuracy of 97.11% on a dataset of 2978 individual text blocks. M. B. Vijayalaxmi *et al.* [7] proposed a script identification of *Roman*, *Devanagari*, *Kannada*, *Tamil*, *Telugu* and *Malayalam* scripts at text block level using features of Correlation property of Gray Level Co-occurrence Matrix (GLCM) and multi resolutionality of Discrete Wavelet Transform (DWT). Using Support Vector Machine (SVM) classifier, the average script classification accuracy achieved in case of bi-script and tri-script combinations were 96.4333% and 93.9833% respectively on a dataset of 600 text block images. M. C. Padma *et al.* [8] presented a texture-based approach to identify the script type of the documents printed in 3 prioritized scripts *namely*, *Kannada*, *Hindi* and *English*. The texture features were extracted from the Shannon entropy values, computed from the sub-bands of the wavelet packet decomposition. Script classification performance was analyzed using *k*-NN classifier on a dataset of 2700 text images and the average success rate was found to be 99.33%. Sk. Md. Obaidullah *et al.* [9] proposed an automatic handwritten script identification technique at block level for document images of six popular *Indic* scripts *namely*, *Bangla*, *Devanagari*, *Malayalam*, *Oriya*, *Roman* and *Urdu*. Initially, a 34-dimensional feature vector was constructed applying different transforms (based on Radon Transform, Discrete Cosine Transform, Fast Fourier Transform and Distance Transform), textural and statistical techniques. Finally, using a GAS (Greedy Attribute Selection) method, 20 attributes were selected for learning process. Experimentation performed on a dataset of 600 image blocks using Logistic Model Tree which produced an accuracy of 84%. It is apparent from the literature survey that only a few works [4-5, 9] have been done on handwritten *Indic* scripts. Estimation of features of individual scripts from the whole document image involves a lot of computation time because of large image size. Apart from this, the major drawback of the existing works is that the recognition of scripts is mainly performed on individual fixed-sized text blocks, which are selected manually from the document images. Again, the size of text blocks has a remarkable impact on the accuracy of the overall system. This has motivated us to design an automatic script identification scheme based on texture based features at block level for 11 officially recognized *Indic* scripts *viz.*, *Bangla*, *Devanagari*, *Gujarati*, *Gurumukhi*, *Kannada*, *Malayalam*, *Oriya*, *Tamil*, *Telugu*, *Urdu* and *Roman*. The block selection from the document images has been done automatically by using quad-tree decomposition approach.

## 2 Proposed Work

The proposed model is inspired by a simple observation that every script defines a finite set of text patterns, each having a distinct visual appearance [10]. Scripts are made up of different shaped patterns in forming the different character sets. Individual text patterns of one script are assembled together to form a meaningful text word, a text-line or a paragraph. This collection of the text patterns of the one script exhibits distinct visual appearance. A uniform block of texts, regardless of the content, may be considered as distinct texture patterns (a block of text as single entity) [10]. This observation implies that one may devise a suitable texture classification algorithm to perform identification of text language. In the proposed model, the texture-based features are extracted from the Gabor wavelets filters designed at multiple scales and multiple orientations. These features are estimated at different levels of the document images decomposed by the quad-tree based approach which is described in the next subsection.

## 2.1 Quad-Tree Decomposition Approach

The quad-tree based decomposition approach is designed by recursively segmenting the image into four equal-sized quadrants in top-down fashion. The decomposition rule can be formally defined as follows: The root node of the tree represents the whole image. This method starts dividing each image into four quadrants if the difference between maximum and minimum value of the inspection image is larger than the decomposition threshold, which means that the image is non-homogeneous. This rule is recursively applied to subdivided images until the image cannot be divided. So, at 1st Level, the whole document image is divided into four equal blocks- $L_1$, $L_2$, $L_3$ and $L_4$. For 2nd level, $L_1$ block is again sub-divided into another four sub-blocks *namely*, $L_{11}$, $L_{12}$, $L_{13}$ and $L_{14}$. Similarly, each of the other blocks i.e., $L_2$, $L_3$ and $L_4$ are also sub-divided into four sub-blocks. Thus, for 2nd level decomposition, a total of 16 blocks is realized. Therefore, the total number of sub-blocks at $l$-th level can be written as $4^l$ where $l$ denotes the level of decomposition. For an image of size $M \times N$, then the size of each sub-block at $l$-th level is $(M/2^l) \times (N/2^l)$. For the present work, the maximal value of $l$ is chosen to be 4. Fig. 1 shows the quad-tree decomposition for a handwritten *Telugu* document image.

## 2.2 Gabor Wavelets

The different wavelet transform functions filter out different range of frequencies (i.e. sub bands). Thus, wavelet is a powerful tool, which decomposes the image into low frequency and high frequency sub-band images. Among various wavelet bases, Gabor functions provide the optimal resolution in both the time (spatial) and frequency domains, and the Gabor wavelet transform seems to be the optimal basis to extract local features. Besides, it has been found to yield distortion tolerance for pattern recognition tasks. The Gabor kernel is a complex sinusoid modulated by a Gaussian envelope. The Gabor wavelets have filter responses similar in shape to the respective fields in the primary visual cortex in mammalian brains [11]. The kernel or mother wavelet in the spatial domain in given by [12]:

$$\psi_j(\vec{x}) = \frac{k_j^2}{\sigma^2} exp\left(-\frac{k_j^2 x^2}{2\sigma^2}\right)\left[exp(j\vec{k_j}\vec{x}) - exp\left(-\frac{\sigma^2}{2}\right)\right] \quad (1)$$

where,

$$\vec{k_j} = \begin{pmatrix} k_v \sin\phi_\mu \\ k_v \cos\phi_\mu \end{pmatrix}, k_v = 2^{-\frac{v+1}{2}\pi}, \phi_\mu = \mu\frac{\pi}{8}, \vec{x} = (x_1, x_2) \ \forall \ x_1, x_2 \in \mathbb{R}^2 \quad (2)$$

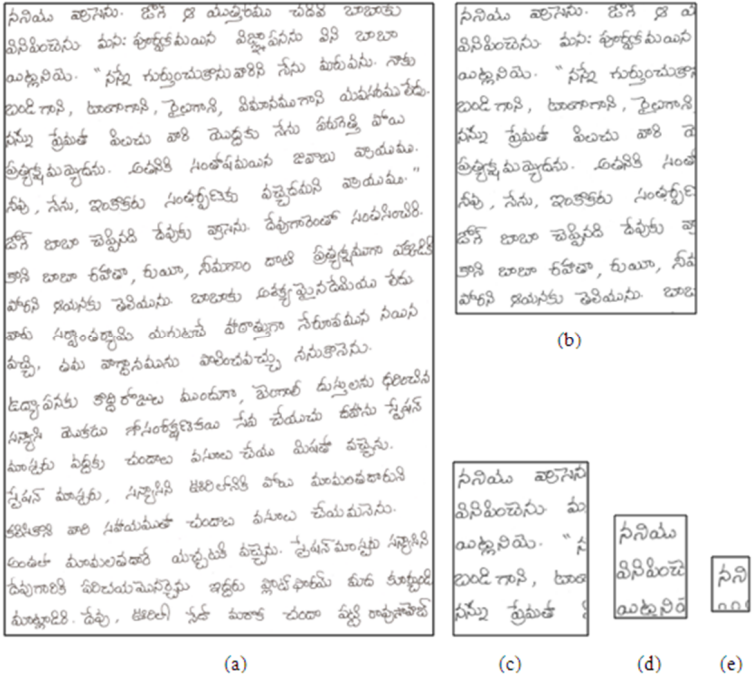

**Figure 1.** Illustration of: (a) handwritten Telugu document page, (b) 1st level, (c) 2nd level ($L_1$), (d) 3rd level ($L_{11}$), and (e) 4th level ($L_{111}$) decomposition

$\sigma$ is the standard deviation of the Gaussian, $\vec{k}$ is the wave vector of the plane wave, $\phi_\mu$ and $k_v$ denote the orientations and frequency scales of Gabor wavelets respectively which are obtained from the mother wavelet. The Gabor wavelet representation of a block image is obtained by doing a convolution between the image and a family of Gabor filters as described by Eqn. (3). The convolution of image $I(x)$ and a Gabor filter $\psi_j(\vec{x})$ can be defined as follows:

$$J_j(\vec{x}) = I(\vec{x}) * \psi_j(\vec{x}) \quad (3)$$

Here, $*$ denotes the convolution operator and $J_j(\vec{x})$ is the Gabor filter response of the image block with orientation $\phi_\mu$ and scale $k_v$. This is referred to as wavelet transform because the family of kernels are self-similar and are generated from one mother wavelet by scaling and rotation of frequency. The transform extracts features oriented along $\phi_\mu$ and for the frequency $k_v$. Each combination of $\mu$ and $v$ results in a sub-band of same dimension as the input image $I$. For the present work, $\mu \in \{0,1,....,5\}$ and $v = \{1,2,.....,5\}$. Five frequency scales and six orientations would yield 30 sub-bands. Fig. 2 shows the directional selectivity property of the Gabor wavelet transform used in the present work whereas Fig. 3 shows the resulting image blocks at five frequency scales and six orientations for a sample handwritten *Kannada* block image obtained at 2nd level of image decomposition. For the feature extraction purpose, we proposed to compute the energy and entropy values [13] from each of the sub-bands of the Gabor wavelet transform which makes the size of our feature vector to 60. Finally, this set of 60 multi-scale and multi-directional oriented features has been used for the block level recognition of 11 official scripts.

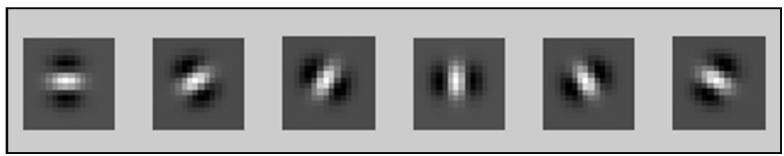

**Figure 2.** Illustration of six kernels of size 16X16 used in the present work

The algorithm of the proposed methodology is described as follows
Step 1: Input a RGB document image and convert it into a gray scale image $I(x, y)$ where, $x$ and $y$ are the spatial coordinates of the image.
Step 2: Select an integer value $l$ where $l$ indicates the level of decomposition. For the present work, the optimal value of $l$ is varied from 2 to 4.
Step 3: Perform $l$-level quad-tree decomposition to generate $4^l$ blocks. Feature values are generated from each of these blocks.
Step 4: Convolve the block image of Step 3 with Gabor wavelet kernel with five different scale values ($v = 1, 2, 3, 4, and\ 5$) and six different orientations ($\phi_\mu = 0^0, 30^0, 60^0, 90^0, 120^0, and\ 150^0$). Thus, for each block we get a total of 30 convolved filter outputs.

Step 5: Acquire the absolute value of these filtered output block images and calculate the energy and entropy values.

Step 6: Repeat Step 2 to Step 5 to extract a total of 60 features from each of the script block images.

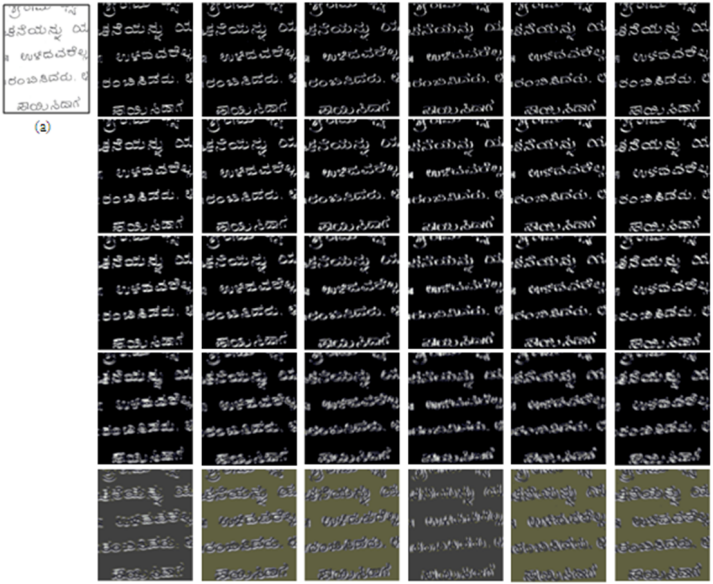

**Figure 3.** Illustration of the output images of Gabor filter responses at five scales and six orientations for a sample handwritten *Kannada* script block image shown in (a). (The first row shows the output for $k_v = 1$ and six orientations, the second row shows the output for $k_v = 2$ and six orientations, and so on)

## 3 Experimental Study and analysis

The experiments are carried out on eleven official scripts namely, *Bangla, Devanagari, Gujarati, Gurumukhi, Kannada, Malayalam, Oriya, Tamil, Telugu, Urdu* and *Roman*. A total dataset of 110 handwritten document pages (10 pages from each script) is collected from different people with varying age, sex, educational qualification etc. These documents are digitized using a HP flatbed scanner and stored as bitmap (BMP) image. Otsu's global thresholding approach [14] is used to convert them into two-tone images (0 and 1) where the label '1' represents the object and '0' represents the background. Binarized block images may contain noisy pixels which have been removed by using Gaussian filter [13]. A set of 160, 640 and 2560 text blocks for each of the scripts are prepared by applying 2-level, 3-level and 4-level quad-tree based page segmentation approach. Sample handwritten text block images of 2-level decomposition for all the above mentioned scripts are shown in Fig. 4. A 3-fold cross-validation approach is used for testing the present approach. That is, for 2-level decomposition approach, a total of 1175 text blocks are used for the training purpose and the remaining 585 text blocks are used for the testing purpose. Similarly, for 3-level and 4-level decomposition approaches, a set of 4695 and 18775 text blocks are taken for the training purpose and the remaining 2345 and 9285 text blocks are used for testing the system respectively. The proposed approach is then evaluated using six different classifiers *namely*, Naïve Bayes, MLP, SVM, Random Forest, Bagging, MultiClass classifier. The graphical comparison of the identification accuracy rates of the said classifiers for 2-level, 3-level and 4-level approaches are shown with the help of a bar chart in Fig. 5.

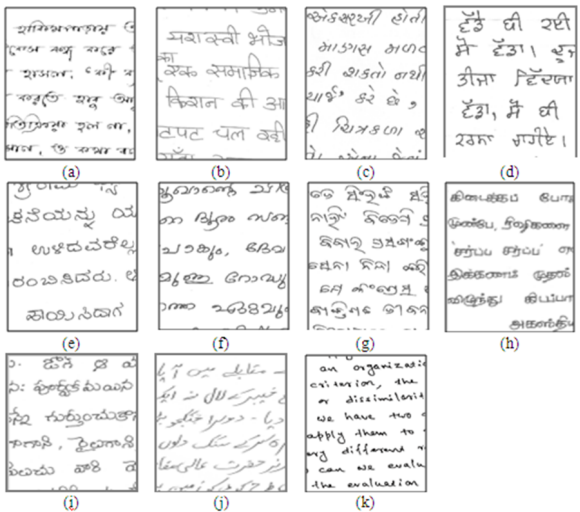

**Figure 4.** Samples of text block images obtained at 2-level quad-tree based decomposition scheme written in: (a) *Bangla*, (b) *Devanagari*, (c) *Gujarati*, (d) *Gurumukhi*, (e) *Kannada*, (f) *Malayalam*, (g) *Oriya*, (h) *Tamil*, (i) *Telugu*, (j) *Urdu*, and (k) *Roman* scripts respectively

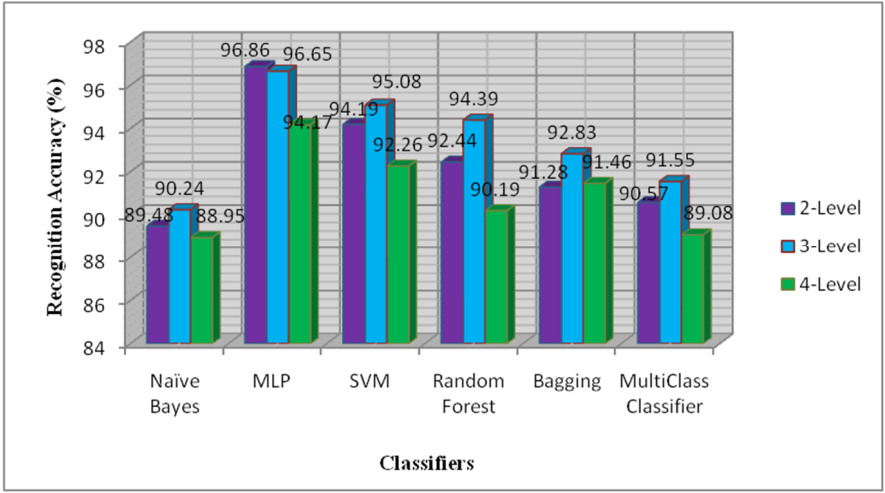

**Figure 5.** Graph showing the recognition accuracies of the proposed script identification technique for 2-level, 3-level and 4-level decomposition approaches using six different classifiers

It can be seen form Fig. 5 that the best identification accuracy is found to be 96.86% for 2-level decomposition approach by MLP classifier. In the present work, detailed error analysis of MLP classifier with respect to different well-known parameters *namely*, Kappa statistics, mean absolute error (MAE), root mean square error (RMSE), True Positive rate (TPR), False Positive rate (FPR), precision, recall, F-measure, and Area Under ROC (AUC) on eleven handwritten scripts are also computed. Table 1 provides a statistical performance analysis of MLP classifier for text blocks written in each of the aforementioned scripts.

Although the recognition accuracy achieved by the present technique is quite impressive considering the inherent complexities of scripts involved, but still some of the block images written in one script are misclassified into another script. The main reasons for this misclassification are due to presence of noise, distortion, and peculiar writing styles of different writers. Apart from this, sometimes the sub-block images (decomposed from the document image) may contain more background part than the foreground ones which in turn estimate the feature values misleading enough for the present script identification technique.

**Table 1.** Statistical performance measures of MLP classifier along with their respective means (shaded in grey and styled in bold) achieved by the present technique on 11 handwritten scripts at block level

| Scripts | Kappa Statistics | MAE | RMSE | TPR | FPR | Precision | Recall | F-measure | AUC |
|---|---|---|---|---|---|---|---|---|---|
| Bangla | | | | 0.938 | 0.008 | 0.938 | 0.938 | 0.938 | 0.983 |
| Devanagari | | | | 0.956 | 0.003 | 0.975 | 0.956 | 0.965 | 0.983 |
| Gujarati | | | | 0.988 | 0.001 | 0.994 | 0.988 | 0.991 | 1.000 |
| Gurumukhi | | | | 1.000 | 0.005 | 0.964 | 1.000 | 0.982 | 0.999 |
| Kannada | | | | 0.981 | 0.001 | 0.994 | 0.981 | 0.987 | 0.996 |
| Malayalam | 0.9647 | 0.1917 | 0.305 | 0.988 | 0.005 | 0.958 | 0.988 | 0.972 | 1.000 |
| Oriya | | | | 0.969 | 0.002 | 0.981 | 0.969 | 0.975 | 0.992 |
| Tamil | | | | 0.938 | 0.003 | 0.974 | 0.938 | 0.955 | 0.988 |
| Telugu | | | | 0.961 | 0.007 | 0.942 | 0.961 | 0.951 | 0.982 |
| Urdu | | | | 0.969 | 0.004 | 0.971 | 0.969 | 0.995 | 1.000 |
| Roman | | | | 0.971 | 0.002 | 0.968 | 0.971 | 0.948 | 0.989 |
| **Mean** | **0.9647** | **0.1917** | **0.305** | **0.969** | **0.004** | **0.969** | **0.969** | **0.969** | **0.992** |

# Conclusion

Script identification, a challenging research problem in any multilingual environment, has got attention to the researchers few decades ago. Research in the field of script identification aims at conceiving and establishing an automatic system which would be able to discriminate a certain number of scripts. As developing a common OCR engine for different scripts is near to impossible, it is necessary to identify the scripts correctly before feeding them to corresponding OCR engine. In this work, we assessed the effectiveness of using Gabor wavelets filters on 11 officially recognized handwritten *Indic* scripts along with *Roman* script. The scheme is applied at block level using a quad-tree based decomposition approach. The present technique attains an identification accuracy of 96.86% for 2-level decomposition. As the present technique is evaluated on a limited dataset, we have achieved satisfactory result. This work is first of its kind to the best of our knowledge as far as the number of scripts is concerned. As the key feature used in this technique is mainly texture based, in future, the technique could be applicable for recognizing other non-*Indic* scripts in any multi-script environment. Future scope may also include comparing different wavelet transforms with the Gabor wavelet transform in terms of computational accuracy.

## Acknowledgment

The authors are thankful to the Center for Microprocessor Application for Training Education and Research (*CMATER*) and Project on Storage Retrieval and Understanding of Video for Multimedia (SRUVM) of Computer Science and Engineering Department, Jadavpur University, for providing infrastructure facilities during progress of the work. The current work, reported here, has been partially funded by Technical Education Quality Improvement Programme Phase–II (TEQIP-II), Jadavpur University, Kolkata, India.

## References

1. H. Scharfe, ''*Kharosti and Brahmi*'', J. Am. Oriental Soc., vol. 122, no. 2, pp. 391-393, 2002.
2. A. S. Mahmud, ''*Crisis and Need: Information and Communication Technology in Development Initiatives Runs through a Paradox*'', ITU Document WSIS/PC-2/CONTR/17-E, World Summit on Information Society, In: International Telecommunication Union (ITU), Geneva, 2003.
3. P. K. Singh, R. Sarkar, M. Nasipuri: "*Offline Script Identification from Multilingual Indic-script Documents: A state-of-the-art*", In: Computer Science Review (Elsevier), vol. 15-16, pp. 1-28, 2015.
4. M. Hangarge, B. V. Dhandra, "*Offline Handwritten Script Identification in Document Images*", In: International Journal of Computer Applications, vol. 4, no. 6, pp. 6-10, 2010.
5. G. G. Rajput, Anita H. B., "*Handwritten Script Recognition using DCT and Wavelet Features at Block Level*", In: IJCA Special Issue on "Recent Trends in Image Processing and Pattern Recognition", pp. 158-163, 2010.
6. G. D. Joshi, S. Garg, J. Sivaswamy, "*Script Identification from Indian documents*", In: DAS, LNCS 3872, pp. 255-267, 2006.
7. M. B. Vijayalaxmi, B. V. Dhandra, "*Script Recognition using GLCM and DWT Features*", In: International Journal of Advanced Research in Computer and Communication Engineering, vol. 4, issue 1, pp. 256-260, 2015.
8. M. C. Padma, P. A. Vijaya, "*Entropy Based Texture Features Useful for Automatic Script Identification*", In: International Journal on Computer Science and Engineering, vol. 2, no. 2, pp. 115-120, 2010.
9. Sk. Md. Obaidullah, C. Halder, N. Das, K. Roy, "*Indic Script Identification from Handwritten Document Images-An Unconstrained Block-level Approach*", In: Proc. of 2[nd] IEEE International Conference on Recent Trends in Information Systems, pp. 213-218, 2015.
10. T. N. Tan, "*Rotation Invariant Texture Features and their use in Automatic Script Identification*", In: IEEE Transactions on Pattern Analysis and Machine Intelligence, vol. 20, no. 7, pp. 751-756, 1998.
11. J. G. Daugman, "*Uncertainty relation for resolution in space, spatial-frequency, and orientation optimized by two-dimensional visual cortical filters*", In: J. Optical Soc. Amer., 2(7), pp. 1160-1169, 1985.

**12** T. S. Lee, "*Image representation using 2D Gabor wavelets*", In: IEEE Transactions on Pattern Analysis and Machine Intelligence, vol. 18, no. 10, 1996.
**13** R. C. Gonzalez, R. E. Woods, "*Digital Image Processing*", vol. I. Prentice-Hall, India (1992).
**14** N. Ostu "*A thresholding selection method from gray-level histogram*", In: IEEE Transactions on Systems Man Cybernetics, SMC-8, pp. 62-66, 1978.